\pdfoutput=1
\documentclass[11pt]{article}

\usepackage[final]{acl}

\usepackage{times}
\usepackage{latexsym}

\usepackage[T1]{fontenc}
\usepackage[utf8]{inputenc}

\usepackage{microtype}

\usepackage{amsmath,amssymb,amsthm}
\usepackage[dvipsnames,svgnames]{xcolor}
\usepackage{cancel}
\newcommand\hcancel[2][black]{\setbox0=\hbox{$#2$}%
\rlap{\raisebox{.45\ht0}{\textcolor{#1}{\rule{\wd0}{1pt}}}}#2}

\usepackage{subcaption}
\usepackage{tablefootnote}
\usepackage{booktabs}
\usepackage{enumitem}
\usepackage{graphicx}
\usepackage{multirow}
\usepackage{bm}
\usepackage{dsfont}
\newcommand{\cites}[1]{\citeauthor{#1}'s \citeyearpar{#1}}

\definecolor{darkgreen}{HTML}{005e19}
\definecolor{darkblue}{HTML}{240394}
\newcommand{\code}[1]{\texttt{#1}}
\newcommand{\exampleg}[1]{\textcolor{darkgreen}{\textbf{\code{#1}}}}

\newcommand{\exampler}[1]{\textcolor{red}{\textbf{\code{#1}}}}

\usepackage{inconsolata}

\usepackage{graphicx}

\title{Sheaf Discovery with Joint Computation Graph Pruning and Flexible Granularity}

\author{
    Lei Yu$^{1}$\thanks{Equal contribution.},\;
    Jingcheng Niu$^{12}$\footnotemark[1],\;
    Zining Zhu$^{13}$,\;
    Xi Chen$^{1}$,\;
    Gerald Penn$^{1}$\\
    $^1$University of Toronto,\;
    $^2$UKP Lab, TU Darmstadt,\;
    $^3$Stevens Institute of Technology\;
}

\begin{document}
\maketitle
\begin{abstract}
  In this paper, we introduce DiscoGP, a novel framework for extracting self-contained modular units, or \textit{sheaves}, within neural language models (LMs). Sheaves extend the concept of functional \textit{circuits}, a unit widely explored in interpretability research, by considering not only subsets of edges in an LM's computation graph but also the model's weight parameters. Our framework identifies sheaves through a gradient-based pruning algorithm that operates on both of these in such a way that reduces the original LM to a sparse skeleton that preserves certain core capabilities. Experimental results demonstrate that, across a range of linguistic and reasoning tasks, DiscoGP extracts sheaves that preserve 93-100\% of a model's performance on the identified task while comprising only 1-7\% of the original weights and connections. Furthermore, our analysis reveals that, compared to previously identified LM circuits, the sheaves discovered by DiscoGP exhibit superior modularity and functional fidelity. Extending our method to the neuron level also unveils novel insights into the inner workings of LLMs.%
  \footnote{The code and results of DiscoGP are available online: \url{https://github.com/frankniujc/disco_gp}.}
\end{abstract}

\section{Introduction}

Systems built with transformer language models \citep[LMs;][]{vaswaniAttentionAllYou2017, devlinBERTPretrainingDeep2019, radfordLanguageModelsAre2019, raffelExploringLimitsTransfer2020, openaiGPT4TechnicalReport2023, touvronLlama2Open2023} have demonstrated incredible capabilities in solving various natural language tasks across different fields.  The exact mechanisms by which these models achieve these results remain poorly understood, however. Researchers in the field of interpretability therefore aim to provide human-understandable explanations of the computational mechanisms of these ``black-boxed'' LMs. Should one of these interpretations become available, it could lead to the improvement of LMs with better controllability and performance, and even germinate a subsequent generation of explainable artificial intelligence (XAI) systems.

\begin{figure}
    \centering
    \includegraphics[width=0.9\linewidth]{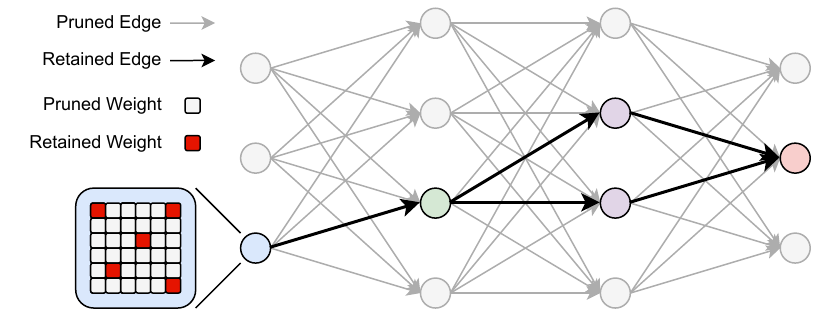}
    \caption{Illustration of DiscoGP. By combining computation-edge and weight-parameter pruning, DiscoGP achieves better performance with neuron-level granularity.}
    \label{fig:open-figure}
    % \vspace{-1em}
\end{figure}

Now, a nascent ``circuit''-based framework has emerged that aims to explain this process and provide the most convincing explanation of LM behaviours to date. This generally decomposes the computation process of an LM into a directed acyclic graph (DAG) and identifies a subset of model components and connections (information flow) that correspond to specific model behaviours. Initially, these circuits were identified manually using various activation or attention patching methods \citep{wangInterpretabilityWildCircuit2022}. ACDC~\citep{conmyAutomatedCircuitDiscovery2023} automated the circuit-discovery process. Since then, several follow-up attempts \citep{syedAttributionPatchingOutperforms2024,hannaHaveFaithFaithfulness2024,zhangEAPGPMitigatingSaturation2025} have been proposed to further advance the state-of-the-art in circuit discovery.

The term \emph{circuit}, however, is used to refer to several distinct concepts, even within the LM interpretability community. We provide a survey of the nomenclature (\S\ref{sec:sheaves_and_circuits}) and clarify our own intentions in respect of interpretation. With a thorough and rigorous definition of the computation graph, edge pruning, and weight pruning, we introduce \emph{sheaf} as a new technical term: a subset of edges in the computation graph and weights in the model that, when executed in isolation, can preserve the original model's behaviour or capabilities on a given task. Simply put, we seek to identify the interpretable sheaf of model components within the LM ``haystack.''

Sheaf discovery fills a gap in current mechanistic interpretability research in that it identifies a self-contained collection of model units that can perform a particular LM function in isolation.
Whereas recent attempts have boasted of their performance on the basis of \textit{post hoc} dissections revealing that the discovered circuit was of crucial importance, our own perspective is that, for all but the most esoteric of purposes, circuits are not worth discovering if they cannot operate by themselves.  For this reason, we follow the regimen of ``zero ablation,'' in which layer heads not identified during training as part of a sheaf are zeroed out during evaluation.  Until now, although zero ablation is occasionally acknowledged as a worthwhile goal, this has not been standard practice.
Sheaves offer a unique opportunity to manipulate self-contained units and gain novel insights into the internal workings of transformer-based LMs.
% In doing so, we can pinpoint the components and connections corresponding to specific model behaviours, providing a granular, faithful, and complete explanation of how the model performs a given task.

Prior automatic circuit-discovery methods share a crucial limitation, moreover: the computational power they require is prohibitively large because the number of edges in the computation graph grows quadratically ($O(n^2)$) with the number of model components. This has prevented researchers from engaging in discovery at the neuronal level.  The presence of a parallel thread of investigation into the properties of MLP neurons, finding them to be highly idiosyncratic in both the type of information they contain and their function within large neural networks \citep{gevaTransformerFeedForwardLayers2021,daiKnowledgeNeuronsPretrained2022,mengLocatingEditingFactual2022,niuWhatDoesKnowledge2024,hongIntrinsicEvaluationUnlearning2024}, however, does suggest that refining the granularity of these interpretations to the neuronal level would be valuable. Recent work on the ``knowledge neuron thesis'' \citep{niuWhatDoesKnowledge2024,daiKnowledgeNeuronsPretrained2022}, for example, has shown that modifying just a few neurons, or even a single neuron, can lead to substantial changes in the model's behaviour.

The \textbf{Dis}covery with Joint \textbf{Co}mputation \textbf{G}raph \textbf{P}runing (DiscoGP) framework addresses the granularity and scaling problem by applying joint computation-edge and model-weight parameter pruning with gradient-based masking. While the computation graph is still defined at the relatively coarse level of attention heads and MLPs, as in other circuit-discovery methods, DiscoGP extends this approach to weight pruning within each individual computation-graph node to enable finer, neuron-level discovery.

DiscoGP achieves state-of-the-art performance in sheaf detection: it identifies the sheaves for a wide range of tasks with the fewest edges and weight parameters while maintaining near-perfect performance compared to the original model's performance. By refining granularity to the neuronal level, we unveil several critical insights into the model that were previously unavailable.

We begin with a formal definition of \textbf{sheaves}, and provide a survey that clarifies the various different uses of the term \emph{circuit} in relevant literature (\S\ref{sec:sheaves_and_circuits}). Then, we introduce \textbf{DiscoGP}, a novel sheaf-discovery framework with joint pruning of weight parameters and computation-graph edges that enables individual neuron-level granularity (\S\ref{sec:disco_gp}). Using DiscoGP, we can obtain sheaves across a wide range of tasks that are \textbf{sparser and yet more faithful} to the original model.

\section{Sheaves and Circuits}
\label{sec:sheaves_and_circuits}

In this section, we present a comprehensive definition of the main task of \textit{sheaf discovery}, and discuss its similarities and differences compared to the broad range of tasks often referred to as ``circuit discovery'' in the literature, as the term is used inconsistently and can sometimes cause confusion.
We start with a survey of the different definitions of circuit discovery (\S\ref{sub:survey}), and then introduce our sheaf-based framework by defining weight pruning (\S\ref{sub:weight_pruning}) and edge pruning (\S\ref{sub:edge_pruning}).

\subsection{Survey: Circuits and Circuit Discovery}
\label{sub:survey}

\paragraph{Circuit}

The term ``circuit'' has various meanings within the LM interpretability community, depending on the context. \citet{nandaComprehensiveMechanisticInterpretability2022} describes it as ``a fairly fuzzy and poorly defined term'' that roughly refers to ``the sub part [sic] of a model that does some understandable computation to produce some interpretable features from prior interpretable features.'' \citet{olahZoomIntroductionCircuits2020} considered circuits as a set of features and the weighted connections between them. \citet{elhageMathematicalFrameworkTransformer2021} used the term ``circuit'' to refer to the separable parts of the computation process within each attention head. Because the computation of a transformer model can generally be considered linear, \citet{elhageMathematicalFrameworkTransformer2021} argued that the computation of the query and key matrices and the output and value matrices can be considered as two largely independent QK and OV circuits. More recently, work in the field typically decomposes an LM into functional ``building blocks'' and considers the collection of these blocks and a subset of their connections as a circuit; but what constitutes building blocks may still differ from paper to paper. \citet{wangInterpretabilityWildCircuit2022} referred to a circuit as the collection of attention heads, while ACDC used the term ``circuit'' to refer to the subset of edges between attention heads and MLPs in the computation graph.

\begin{figure*}[t]
  \begin{subfigure}{0.32\linewidth}
    \centering
    \includegraphics[width=\linewidth]{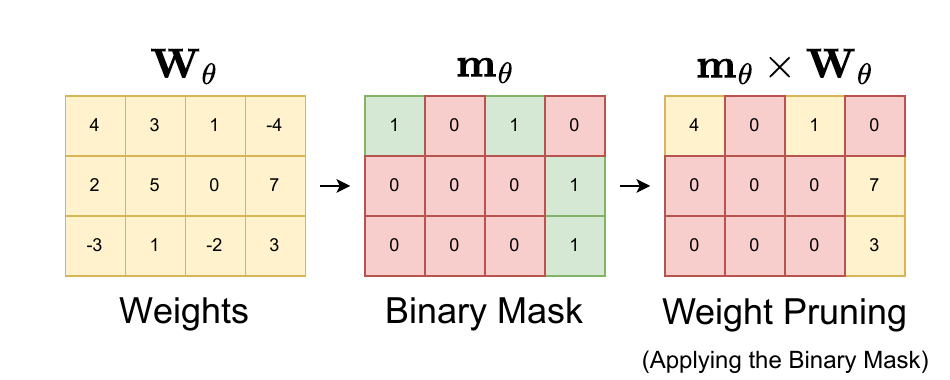}
    % \vspace{-1em}
    \caption{Weight pruning is performed by directly setting the value of a weight parameter to zero. In practice, previous work has shown that gradient-based masking is one of the most effective methods for weight pruning. Specifically, a binary mask is applied to each weight parameter, and the algorithm optimises the values of these masks to identify the subnetwork that achieves the best task performance.}
    \label{fig:weight_pruning}
  \end{subfigure} \hfill
  \begin{subfigure}{0.28\linewidth}
    \centering
    \includegraphics[width=0.9\linewidth]{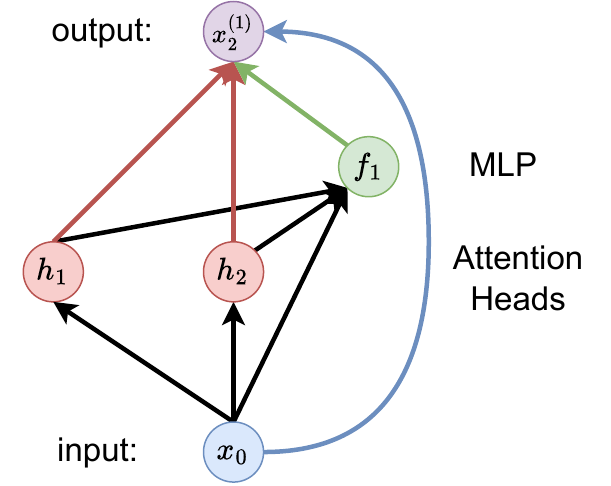}
    \vspace{-0.5em}
    \caption{The computation graph of the single-layer transformer block example. This graphical representation corresponds to the unrolling of the residual stream in (\ref{eq:unroll_1_layer}). The top-level terms are colour-coded to match those in (\ref{eq:unroll_1_layer}).}
    \label{fig:small_comp_graph}
  \end{subfigure} \hfill
  % \begin{subfigure}{0.3\linewidth}
  % BIGGER COMP GRAPH
  % \caption{The computation graph of a complete transformer model with 2 layers and 2 attention heads. The residual stream unroll is presented in Appendix \ref{app:computation_graph_example}.}
  % \label{fig:big_comp_graph}
  % \end{subfigure}\hfill
  \begin{subfigure}{0.37\linewidth}
    \centering
    \begin{minipage}[b]{0.4\linewidth}
        \includegraphics[width=\linewidth]{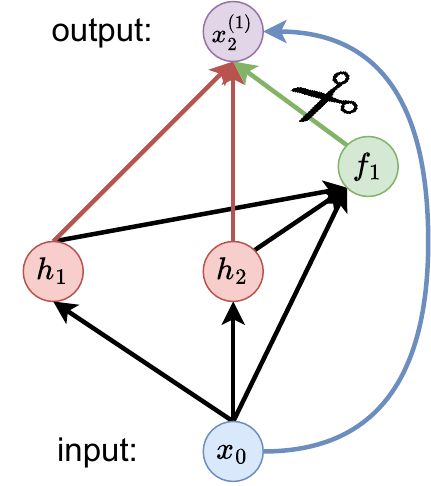}
        \vspace{-2em}
    \end{minipage}\hfill
    \begin{minipage}[b]{0.52\linewidth}
    {\small Cutting off the edge from MLP to output:}\\
    $x_1 = \textcolor{blue}{x_0} + \textcolor{red}{h_1(x_0)} $ $ + \textcolor{red}{h_2(x_0)} 
      \displaystyle + \hcancel[red]{
      \displaystyle \textcolor{DarkGreen}{f_1(x_1^{\text{mid}})}}$
    \end{minipage}
    \caption{``Cutting off'' an edge has somehow become ambiguous.  In the zero-ablation setting, it means removing a term from the residual stream entirely. In other settings (mean- or interchange-ablation), the term is merely replaced with a disingenuous value.  Circuits in the latter view are validated by perturbing the values of those terms and showing that performance of the resulting network is largely invariant.}
    \label{fig:pruning}
  \end{subfigure}
  \caption{Illustration of the formulation of sheaf discovery: computation graph, weight pruning and edge pruning.}
  % \vspace{-1em}
\end{figure*}

\paragraph{Circuit Discovery}

The task of identifying the aforementioned circuits in pre-trained transformer LMs is called \emph{circuit discovery}. Early studies typically searched for circuits manually during simple tasks such as rudimentary anaphora resolution \citep{wangInterpretabilityWildCircuit2022} or simple arithmetic reasoning \citep{hannaHowDoesGPT22023}, using a combination of interpretability tools, including causal interventions \citep{vigInvestigatingGenderBias2020,mengLocatingEditingFactual2022} and logit lenses \citep{gevaTransformerFeedForwardLayers2022,gevaDissectingRecallFactual2023,yuMechanisticUnderstandingMitigation2024}. More recently, ACDC \citep{conmyAutomatedCircuitDiscovery2023} automated the circuit-discovery process. Specifically, they used the activation-patching technique \citep{goldowsky-dillLocalizingModelBehavior2023,zhangBestPracticesActivation2023}, or its approximations \citep{nandaAttributionPatchingActivation2023}, to assess a computation edge's importance by first knocking it out and observing its effect on the model's final output. Beginning at the output node and proceeding in reverse topological order, they evaluate the effect of removing each of the node's incoming edges individually. If the removal of an edge has a greater effect than a predetermined threshold, the edge is included in the circuit; otherwise, it is pruned. \citet{syedAttributionPatchingOutperforms2024} extended ACDC with attribution patching to achieve improved results.

Recent work has also explored other notions of circuithood, such as formulating circuits as collections of human-interpretable neural activation features \citep{hubenSparseAutoencodersFind2024,marksSparseFeatureCircuits2024,yuRobustLLMSafeguarding2024}, collections of attention heads \citep{niuLlamaSeeLlama2025}, or as distributed neural representations of proposed symbolic algorithms \citep{geigerCausalAbstractionsNeural2021,wuInterpretabilityScaleIdentifying2023}.

Most automated circuit-discovery studies evaluate their methods based on their structural overlap with previously discovered or manually hardwired circuits \citep{conmyAutomatedCircuitDiscovery2023,syedAttributionPatchingOutperforms2024}. We concur with recent critiques of this evaluation metric \citep{hannaHaveFaithFaithfulness2024}, and note that the functional fidelity (often quaintly termed functional \emph{faithfulness}) metric, measuring how well the circuit reproduces the original model's performance, is a more appropriate criterion for this task.

\subsection{Weight Pruning}
\label{sub:weight_pruning}

Weight pruning is a technique widely used in the model interpretability community to identify subnetworks (a subset of a model's weight parameters) associated with specific functions of a neural network \citep{caoLowComplexityProbingFinding2021,csordasAreNeuralNets2021,zhangCanSubnetworkStructure2021,guoParameterEfficientTransferLearning2021,decaoSparseInterventionsLanguage2022}. More recently, \citet{leporiBreakItEvidence2023} extended this work to transformer-based language models. Figure~\ref{fig:weight_pruning} provides an overview of weight pruning. This line of research has been encouraged in part by \cites{frankleLotteryTicketHypothesis2019} Lottery Ticket Hypothesis, which states that it is possible to identify smaller functional subnetworks even within dense, randomly initialized models. When this subnetwork is trained from scratch with a similar computational budget, it can achieve performance comparable to that of the original model. Using the \emph{continuous sparsification} method (Figure \ref{fig:weight_pruning}), \citet{savareseWinningLotteryContinuous2020} demonstrated that this subnetwork can be directly extracted from a neural network that maintains task performance without the need for retraining, as originally suggested in the hypothesis. The method is also sometimes referred to as \emph{differentiable masking} \citep{decaoSparseInterventionsLanguage2022}.  It is not always --- there are many ways to approximate a gradient without using analytic differentiation.

\subsection{Computation Graph and Edge Pruning}
\label{sub:edge_pruning}

% Recent research in mechanistic interpretability suggests that investigating weight parameters and activations may not be the most effective approach to understanding the internal workings of LLMs, as several critical limitations of weight- and activation-based approaches have been identified \citep{haseDoesLocalizationInform2023,niuWhatDoesKnowledge2024}. In contrast, \emph{circuit discovery}, which investigates the flow of information within the network, has shown more promising results in revealing how LLMs function, offering a more comprehensive understanding than weight-based analysis. With refinements to circuit searching methods and improvements in functional fidelity, circuit discovery could enable better decomposition of language models. Moreover, weight and edge pruning are not mutually exclusive; why not use both? Our DiscoGP framework integrates the two approaches for more degrees of freedom and achieves the state-of-the-art result in LM decomposition (\S\ref{sec:disco_gp}).

\paragraph{Computation Graph}
\citet{elhageMathematicalFrameworkTransformer2021} introduced the concept of a \textit{residual stream}, providing a clear and concise view of the computation within a transformer block. Each block  consists of an attention module followed by an MLP module. Encoder blocks are stacked like layers of a neural network, but they have many layers inside them.  Let $x_i$ be the input to the $i$-th transformer block, with $H^{(i)}$ representing the set of attention heads and $f_i$ denoting the MLP module, we can write the output of the $i$-th block ($x_{i+1}$) in a transformer as:
\begin{equation}
x_{i+1} = \overbrace{x_{i} + \sum_{h\in{H^{(i)}}} h(x_i)}^{x_i^\text{mid}} + f_i(x_i^\text{mid}) .
\end{equation}
To demonstrate the concept of a computation graph, let us consider a simple one-block transformer model with the input embedding $x_0$. We can ``unroll'' the residual stream \citep{nandaTransformerLens2022} as shown in (\ref{eq:unroll_1_layer}). From this stream, we can tell that the final output of this small transformer consists of 4 terms: the original word embedding input $\textcolor{blue}{x_0}$, the output of the two attention heads $\textcolor{red}{h_1(x_0)}$ and $\textcolor{red}{h_2(x_0)}$, as well as the output of the MLP module $\textcolor{DarkGreen}{f_1(x_0 + h_1(x_0) + h_2(x_0))}$.
\begin{equation}
\begin{array}{lcl}
x^{\text{mid}}_1 & = & x_0 + h_1(x_0) + h_2(x_0) \\
x_1 & = & x^{\text{mid}}_1 + f_1(x^{\text{mid}}_1) = \textcolor{blue}{x_0} + \textcolor{red}{h_1(x_0)} +  \\
& & \textcolor{red}{h_2(x_0)} + \textcolor{DarkGreen}{f_1(x_0 + h_1(x_0) + h_2(x_0))}.
\end{array}
\label{eq:unroll_1_layer}
\end{equation}
From the unrolled residual stream, we can understand how information flows within the transformer block. Using (\ref{eq:unroll_1_layer}) as an example, the output ($x_2$) is derived from the outputs of the two attention heads ($h_1$, $h_2$), the MLP ($f_1$) output, and the original input embedding ($x_0$). The attention heads only take $x_0$ as input, while $f_1$ receives both the outputs of the attention heads and $x_0$. Based on this information flow, we can construct a computation graph as shown in Figure \ref{fig:small_comp_graph}.

\paragraph{Edge Pruning}
\label{par:edge_pruning}

The introduction of computation graphs allows us to analyse the impact of information flow between model components.
Examining how pruning\footnote{Also referred to as {\it knockout}, {\it cut-off}, or {\it ablation}.} a computation edge in this graph affects the model's final output reveals the importance of that specific information flow. In DiscoGP, as shown in Figure~\ref{fig:pruning}, the pruning of an edge is equivalent to the removal of a term in the unrolled residual stream, which can be achieved either through greedily applying causal mediation methods \citep{vigInvestigatingGenderBias2020,finlaysonCausalAnalysisSyntactic2021,mengLocatingEditingFactual2022} to identify important edges \citep{conmyAutomatedCircuitDiscovery2023}, or by leveraging gradient-based techniques to mask out non-essential component connections \citep{bhaskarFindingTransformerCircuits2024}. 

% To model a computational graph without particular edge connections, the most straightforward way is to directly set its transmitted activation value to zero (\citet{conmyAutomatedCircuitDiscovery2023} coined the term {\it zero ablation}). Yet, zero is not the only ablation option. \citet{conmyAutomatedCircuitDiscovery2023} introduced two additional ``patching''-based strategies: \emph{mean ablation}, which sets the activation to the average output value across a reference distribution obtained by running a sample dataset through the model; and \emph{interchange ablation}, which replaces the activation with its value from a corrupted input, obtained by modifying specific input tokens. Both patching strategies involve substituting the model's activations with cached values from previous runs or averaged results. We argue that, therefore, zero ablation is the most suitable method for modularizing LLMs using circuits, as the other two approaches require caching the entire model's activation histories.

\section{DiscoGP: Sheaf Discovery}
\label{sec:disco_gp}

Weight pruning and circuit pruning are not mutually exclusive, so why not apply both? Here, we introduce the term ``sheaf'' to describe the intersection of our particular brand of circuit pruning (edge pruning) and subnetwork pruning (weight pruning). Let $G = \{E, V\}$ represent the computation graph, and let $\bm{\Theta}$ denote the set of all parameters of the model. The task of identifying a sheaf involves searching for two binary masks, $\bm{m} = (\bm{m}_{\bm{\theta}}, \bm{m}_{E}) \in \{0,1\}^{|\bm{\theta}| + |E|}$, which correspond to the pruned weights and edges, respectively. Similar to prior weight pruning approaches, DiscoGP uses Gumbel sigmoid distributions in both masks, enabling the search for a globally optimal solution across weight and edge pruning. This section outlines the sheaf-discovery procedure and the DiscoGP joint pruning algorithm.

\begin{table*}[t!]
  \centering\small
  \begin{tabular}{l|l|c|c}
  \toprule
  Dataset & \multicolumn{1}{c|}{Example Prompt} & Correct & Incorrect \\
  \midrule
  BLiMP & \exampleg{Raymond is selling this \rule{1em}{0.5pt}} & \exampleg{sketch} & \exampler{sketches} \\
  \midrule
  IOI & \exampleg{When Mary and John went to the store, John gave a drink to \rule{1em}{0.5pt}} & \exampleg{Mary} & \exampler{John} \\
  \midrule
  OQA & \exampleg{The capital city of Canada is \rule{1em}{0.5pt}} & \exampleg{Ottawa} & *not unique \\
  \bottomrule
  \end{tabular}
  \caption{An overview of the tasks and datasets.}
  \label{tab:data}
\end{table*}

In summary, sheaf discovery has three steps:
\begin{enumerate}[leftmargin=*, itemsep=1pt, topsep=1pt]
    \item For an LM capability, define a task corresponding to the capability by constructing a dataset;
    \item Search for a sheaf (a collection of edges and weight parameters) corresponding to the dataset;
    \item Evaluate the functional fidelity of the sheaf -- i.e., determine whether the model can still perform the task after turning off all other components that do not belong to the sheaf.
\end{enumerate}
These three steps of sheaf discovery bear some superficial resemblance to the three steps of the ``automatic circuit-discovery workflow'' proposed by \citet{conmyAutomatedCircuitDiscovery2023}. They argue that researchers should ``perform an extensive and iterative series of patching experiments with the goal of removing as many unnecessary components and connections from the model as possible'' \citep{conmyAutomatedCircuitDiscovery2023}. Our framework differs in two key respects: (1) we do not impose any restrictions on patching-based approaches; and (2) we aim to identify sheaves of high functional fidelity --- in other words, while ACDC's goal is to identify the most salient components and edges, it does not guarantee that the resulting ``circuits'' can perform the task by itself.

\subsection{Joint Weight and Edge Pruning}
Similar to previous work on gradient-based mask learning \citep{louizosLearningSparseNeural2018,csordasAreNeuralNets2021,caoLowComplexityProbingFinding2021,decaoSparseInterventionsLanguage2022,bayazitDiscoveringKnowledgeCriticalSubnetworks2023}, DiscoGP models each mask $m_i \in \bm{m}$ as a random variable, parameterised by a hard-concrete or gumbel-sigmoid distribution. We first compute a continuous score $s_i \in [0,1]$:
\begin{equation}
    s_i = \sigma\Big(\frac{l_i - \log \frac{\log \mathcal{U}_1}{\log \mathcal{U}_2}}{\tau}  \Big),
\end{equation}
where $\tau \in (0, \inf)$ is a temperature hyperparameter, $l_i$ is a learnable logit parameter of a sigmoid distribution $\sigma(\cdot)$, and $\mathcal{U}_1, \mathcal{U}_2 \sim \text{Uniform}(0,1)$ are random variables drawn from a uniform distribution. We then use the straight-through estimator \cite{bengioEstimatingPropagatingGradients2013} to convert the sampled $s_i$ into a binary mask variable:
\begin{equation}
    m_i = [\mathds{1}_{s_i > 0.5} - s_i]_\text{detach} + s_i,
\end{equation}
where $\mathds{1}$ represents the indicator function, and $[\cdot]_\text{detach}$ is an operator that blocks gradient flow during backpropagation. This approach makes the binary mask $m_i$ a differentiable function of the logit $l_i$, allowing it to be optimised through backpropagation for specific objectives.
% We can therefore implement the SP baseline as special cases of DiscoGP by setting $m=1$ for all $m\in\bm{m}_{E}$. To implement HISP, we can simply force all weight masks $\bm{m}_v$ of a node $v\in V$ to have the same value. 

% \paragraph{Residual Unroll and Split QKV}

\paragraph{Sheaf Searching Objectives} Given a task dataset $\mathcal{D} = \{\bf{x}, \hat{\bm{y}}\}$, where $\bf{x}$ represents the input and $\hat{\bm{y}}$ is the output of the original model, our aim is to identify a set of masks $\bm{m}$ on weights and edges, such that the pruned sheaf produces results as close to the original model as possible. To achieve this, we define \textbf{functional fidelity loss} as the negative log-likelihood of the original model’s predicted label in the output distribution of the pruned circuit:
\begin{equation}
    \mathcal{L}_\text{fidelity} = -\sum _{i} \log p_{\bm{m}}(\hat{y}_i|x_i).
\end{equation}
Moreover, we want the sheaf to contain as many function-specific weights and edges as possible. In other words, when the detected sheaf is removed from the original model, the remaining computational graph should perform at near-random levels on ${D}$.
Let $\tilde{\bm{m}} = 1-\bm{m}$ denote the reverse mask of $\bm{m}$, and the \textbf{complementary sheaf} be the sheaf induced by the reverse mask of $\bm{m}$.
We define the \textbf{completeness loss} as the cross-entropy between the output distribution of the complementary sheaf and a uniform distribution over the label space $\{y_k\}_{k=1}^{K}$:
\begin{equation}
    \mathcal{L}_\text{complete} = -\sum _{i}\sum _{k=1}^{K}\frac{1}{K}\log p_{\tilde{\bm{m}}}(y_k|x_i).
\end{equation}
Lastly, we want the sheaf to be as sparse as possible. Therefore, we minimize the \textbf{sparsity loss}:
\begin{equation}
\begin{aligned}
    & \mathcal{L}_\text{sparse} = \mathcal{L}_{\text{sparse}/\bm{\theta}} + \mathcal{L}_{\text{sparse}/E} \\
    & = \frac{1}{|\bm{m}_{\bm{\theta}}|} \sum _{i=1}^{|\bm{m}_{\bm{\theta}}|} \sigma(l_i) + \frac{1}{|\bm{m}_{E}|} \sum _{i=1}^{|\bm{m}_{E}|} \sigma(l_i).
\end{aligned}
\end{equation}
The final objective function is then comprised of a weighted mixture of the three loss terms: 
\begin{align}
    \mathcal{L}_\text{GP} = \mathcal{L}_\text{fidelity} + \lambda_c \mathcal{L}_\text{complete} + \lambda_s \mathcal{L}_\text{sparse},
    \label{eq:DiscoGP-obj}
\end{align}
where $\lambda_c, \lambda_s$ are hyperparameters that regulate relative loss importance. 

\paragraph{DiscoGP Implementation Details} Due to page limitations, other optimisation techniques we implemented, including \textit{post hoc} sheaf pruning and split QKV pruning, are treated in Appendix~\ref{app:implementation_details}.

% \subsection{Discussion: Sheaf Discovery and Automatic Circuit Discovery}

% Here, we highlight the key differences between our proposed task of sheaf discovery and the automatic circuit discovery task explored by \citet{conmyAutomatedCircuitDiscovery2023, syedAttributionPatchingOutperforms2024, bhaskarFindingTransformerCircuits2024}. We have briefly discussed differences in the formulation of the core procedures at the beginning of this section.

% First and foremost, the two tasks differ in their goals and motivations. Let us revisit the famous example studied by \citet{wangInterpretabilityWildCircuit2022}: ``\exampleg{\small When Mary and John went to the store, John gave a drink to \rule{1em}{0.5pt}}'', where \exampleg{\small Mary} is the correct answer and \exampler{\small John} is the incorrect one. The automatic circuit discovery task aims to identify all the important connection edges and components that, when perturbed, cause the greatest change to the final output, and potentially steering the model away from responding \exampleg{\small Mary} to \exampler{\small John}. As we will later show in Section~\ref{sec:experiment_results}, taking these important components alone does not result in a self-contained mechanism that perform the original task.

% The automatic circuit discovery task is interested in identifying the components that, upon perturbed, can lead to the model generating \exampler{\small John} instead of \exampleg{\small Mary}.

% \paragraph{Pruning and Ablating}

\section{Experimental Setup}

\begin{table*}[t!]
    \centering\scriptsize
    \setlength\tabcolsep{3pt}
    \resizebox{\linewidth}{!}{
    \begin{tabular}{l|c|ccc|ccc}
    \toprule

    \multirow{2}{*}{Task}
    & Discovery & Sheaf Acc. (\%) & KL Div. & Comp. Acc. (\%) & Weight Density (\%) & Edge Density (\%) \\
    & Method & {\scriptsize (higher is better)} & {\scriptsize (lower is better)} & {\scriptsize (random$^*$ is better)} & {\scriptsize (lower is better)} & {\scriptsize (lower is better)} \\

    % \midrule
    % \multicolumn{7}{c}{\it BLiMP Syntactic Agreement (BLiMP)} \\
    % \midrule
    \midrule

    & ACDC & 83.3 & 0.121 & 42.7 & 100 & 6.48 \\
    & EAP & 89.3 & 0.091 & 53.9 & 100 & 4.88\\
    anaphor gender 
    & Edge Pruning & 88.4 & 0.137 & 49.7 & 100 & 6.62 \\
    agr. (AGA) & Weight Pruning & 97.1 & 0.078 & 50.2 & 3.01 & 100 \\
    & DiscoGP (Ours) & \textbf{98.5} & \textbf{0.074} & 49.9 & \textbf{1.58} & \textbf{3.88} \\ \midrule

    & ACDC & 81.0 & 0.250 & 67.0 & 100 & 6.26 \\
    & EAP & 95.3 & 0.049 & 56.3 & 100 & 8.66 \\
    anaphor number 
    & Edge Pruning & 87.9 & 0.178 & 39.3 & 100 & 2.78 \\ 
    agr. (ANA) & Weight Pruning & 97.7 & 0.076 & 40.3 & 2.79 & 100 \\ 
    & DiscoGP (Ours) & \textbf{99.7} & \textbf{0.043} & 39.2 & \textbf{1.36} & \textbf{1.94} \\ \midrule

    & ACDC & 85.3 & 0.129 & 46.3 & 100 & 7.35 \\
    & EAP & 85.7 & 0.138 & 40.6 & 100 & 9.83 \\
    det. noun agr. 1
    & Edge Pruning & 83.7 & 0.114 & 59.3 & 100 & 2.27 \\
    (DNA) & Weight Pruning & 95.3 & 0.099 & 53.0 & 0.280 & 100 \\
    & DiscoGP (Ours) & \textbf{95.3} & \textbf{0.098} & 51.7 & \textbf{0.187} & \textbf{1.92} \\ \midrule

    & ACDC & 62.7 & 0.419 & 39.3 & 100 & 6.61\\
    & EAP & 60.0 & 0.434 & 38.3 & 100 & 8.92 \\
    det. noun irr. 1
    & Edge Pruning & 67.1 & 0.374 & 48.0 & 100 & 2.46 \\
    (DNA i) & Weight Pruning & 94.3 & 0.103 & 53.6 & 0.263 & 100 \\
    & DiscoGP (Ours) & \textbf{95.8} & \textbf{0.102} & 47.2 & \textbf{0.244} & \textbf{1.68} \\ \midrule

    & ACDC & 82.4 & 0.169 & 52.3 & 100 & 7.04 \\
    & EAP & 83.5 & 0.153 & 45.7 & 100 & 9.90\\
    det. noun adj. 1
    & Edge Pruning & 50.3 & 0.412 & 47.6 & 100 & 7.14 \\
    (DNA a) & Weight Pruning & 94.7 & 0.136 & 49.9 & 0.565 & 100 \\
    & DiscoGP (Ours) & \textbf{95.5} & \textbf{0.118} & 45.3 & \textbf{0.520} & \textbf{5.71} \\ \midrule

    & ACDC & 50.2 & 0.120 & 41.4 & 100 & 9.46 \\
    & EAP & 60.7 & 0.128 & 44.7 & 100 & 6.89 \\
    det. noun adj. 
    & Edge Pruning & 56.3 & 0.348 & 47.8 & 100 & 12.9 \\
    irr. 1 (DNA ia) & Weight Pruning & 94.6 & 0.127 & 49.9 & 0.569 & 100 \\
    & DiscoGP (Ours) & \textbf{95.1} & \textbf{0.118} & 45.3 & \textbf{0.496} & \textbf{6.22} \\ 

    % \midrule
    % \multicolumn{7}{c}{\it Indirect Object Identification (IOI)} \\
    % \midrule
    \midrule\midrule

    & ACDC & 51.6 & 0.730 & 50.6 & 100 & 2.45 \\
    & EAP & 58.3 & 0.756 & 55.2 & 100 & 3.48 \\
    IOI
    & Edge Pruning & 100 & 0.032 & 49.9 & 100 & 2.97 \\
    & Weight Pruning & 98.4 & 0.043 & 57.5 & 1.87 & 100 \\
    & DiscoGP (Ours) & \textbf{100} & \textbf{0.020} & 49.2 & \textbf{1.79} & \textbf{2.03} \\

    % \midrule
    % \multicolumn{7}{c}{\it PARAREL Open-Domain Question Answering Tasks (PARAREL)$^\dagger$} \\
    % \midrule
    \midrule\midrule

    & ACDC & 1.0 & 0.379 & 0.6 & 100 & 5.35 \\
    & EAP & 0.9 & 0.341 & 0.6 & 100 & 5.92 \\
    PARAREL & Edge Pruning & 90.4 & 0.039 & 0.7 & 100 & 2.97 \\
    Average$^\dagger$ & Weight Pruning & 91.8 & 0.032 & 0.8 & 2.83 & 100 \\
    & DiscoGP (Ours) & \textbf{93.1} & \textbf{0.023} & 0.62 & \textbf{2.77} & \textbf{2.91} \\

    \bottomrule
    \end{tabular}}
    \caption{Sheaf-Discovery Performance Comparison. DiscoGP achieves the best performance across all tasks, using the fewest weight parameters and edges. The best-performing methods are highlighted in {\bf bold}. $^*$: For complement sheaf accuracy, successful searches are expected to yield random performance. Scores close to random therefore indicate good performance, although a direct comparison of complement scores would not be meaningful. BLiMP and IOI’s expected random performance is 50\%, and {\sc ParaRel}'s expected random performance is close to 0\%. $^\dagger$: Due to page limits, only the average training set performance is shown. Full {\sc ParaRel} results are provided in Appendix~\ref{app:pararel_results} and support the same findings.}
    \label{table:main-circuit-results}
    % \vspace{-1em}
\end{table*}

\textbf{Evaluation:}
We evaluate DiscoGP and the baselines across three tasks (Table~\ref{tab:data}): syntactic agreement from the {\bf BLiMP} corpus \citep{warstadtBLiMPBenchmarkLinguistic2020}, the indirect object identification ({\bf IOI}) task introduced by \citet{wangInterpretabilityWildCircuit2022}, and factual information from open-domain question answering (OQA) with the {\bf PARAREL} \citep{elazarMeasuringImprovingConsistency2021} dataset. This ensemble of tasks is intended to provide comprehensive coverage of syntactic, semantic and factual aspects. See Appendix~\ref{app:data} for more information.

Metric-wise, we report the \textbf{functional fidelity}: this includes the sheaf's accuracy and the KL divergence of the sheaf's output (sheaf accuracy refers to the task accuracy when all pruned components are zero-ablated and sheaf KL divergence is measured between the sheaf's output and that of the original model). We also report completeness or the \textbf{complement sheaf accuracy} (i.e., the accuracy when the sheaf is zero-ablated and all other model components are kept on), as well as \textbf{sparsity} (both edge and weight sparsity). These evaluation metrics follow the typical fidelity, completeness and sparsity scheme used by other mechanistic interpretability work \citep{wangInterpretabilityWildCircuit2022, conmyAutomatedCircuitDiscovery2023, syedAttributionPatchingOutperforms2024, bhaskarFindingTransformerCircuits2024}.

\paragraph{Baseline Methods:}
We compare DiscoGP with every other major circuit-discovery method. We categorize the methods into (1) threshold-based greedy search algorithms, which include ACDC~\citep{conmyAutomatedCircuitDiscovery2023} and EAP~\citep{syedAttributionPatchingOutperforms2024}; and (2) gradient-masking-based algorithms, including  weight pruning (WP) methods \citep{louizosLearningSparseNeural2018,caoLowComplexityProbingFinding2021,sanhDistilBERTDistilledVersion2020,decaoSparseInterventionsLanguage2022}, an edge pruning (EP) method~\citep{bhaskarFindingTransformerCircuits2024}, and our novel joint pruning method. See Appendix~\ref{app:baseline} for details about our reproduction of these.

\paragraph{LM Selection:}
We compare to other methods using GPT-2 base (small) model, as it is the only model supported by the original implementation of every method.

% (1) the weight pruning (WP) method learn a binary mask for every model weight parameter, without removing circuit edge connections \citep{louizosLearningSparseNeural2018,caoLowComplexityProbingFinding2021,sanhDistilBERTDistilledVersion2020,decaoSparseInterventionsLanguage2022};  (2) the Automated Circuit Discovery \cite[ACDC;][]{conmyAutomatedCircuitDiscovery2023}, which is a greedy Edge Pruning method that iteratively apply causal attribution patching on node connections \citep{hannaHowDoesGPT22023,goldowsky-dillLocalizingModelBehavior2023}, and (3) the gradient-based edge pruning (EP) method that searches and removes unnecessary edges via differentiable masking \citep{bhaskarFindingTransformerCircuits2024}. Detailed explanations of baseline methods can be found in Appendix \ref{appendix:baselines}.

% We then further evaluate DiscoGP on larger, more modern LLMs including GPT2-XL and Llama-3.2-3B~\citep{grattafioriLlama3Herd2024}.

% We use GPT-2 small as our language model for sheaf discovery, as it has been extensively studied by the mechanistic interpretability research community for similar tasks. For each task, we ran DiscoGP to identify weight and edge mask logits that optimise the learning objective defined in Equation \ref{eq:DiscoGP-obj}, using carefully selected task-specific hyperparameters (see Appendix \ref{appendix:experiment}). 

% Recent work 

\section{Experiment Results}
\label{sec:experiment_results}

% \footnotetext{Closer to random is better}

% \subsection{DiscoGP Outperforms All Baselines}

Table~\ref{table:main-circuit-results} shows the results of DiscoGP compared to the baseline methods. Due to page limits, full results for the OQA task are shown in Appendix~\ref{app:pararel_results}; the breakdown supports the same findings. For each experiment, we run the sheaf-discovery method five times and report average performance. GPT-2 achieves near-perfect performance on all BLiMP and IOI tasks, so we conduct our experiments on the full datasets. GPT-2 performs worse on the OQA PARAREL tasks, on the other hand, so we run experiments only on data samples where the original model answers the question correctly, discarding prompts where it fails, as it is unclear whether searching for a sheaf over a function the LM does not mimic would yield meaningful results.

Overall, we find that DiscoGP outperforms all of the baseline discovery methods. It achieves the highest functional fidelity --- measured either as task accuracy or as KL divergence --- compared to other baselines while using the fewest weight parameters or computation edges. 

\begin{table}[]
    \centering\scriptsize
    \setlength\tabcolsep{1.5pt}
    \begin{tabular}{lccccc}
    \toprule
        \multirow{2}{*}{Method} & Sheaf  & KL Div & Comp. & Weight & Edge \\
         & Acc. (\%) & Acc. (\%) & Acc. (\%) & Density (\%) & Density (\%) \\
    \midrule
        DiscoGP-MA & 100 & 0.014 & 45.8 & 2.14 & 2.49 \\
        ACDC-MA & 51.6 & 0.120 & 41.4 & 100 & 2.45 \\
    \bottomrule
    \end{tabular}
    \caption{When mean ablation (MA) is applied, DiscoGP shows the same kind of performance advantage compared to ACDC.}
    \label{tab:ma}
\end{table}

That said, our method is applicable to other ablation types such as mean ablation, by setting the residual term to the averaged distribution when the mask is 0, instead of setting it to 0. Table~\ref{tab:ma} presents a comparison of DiscoGP-MA to ACDC-MA for the IOI task.

% \begin{table}
% \centering\small
% \begin{tabular}{@{}lccc@{}}
% IOI 
% \end{tabular}
% \caption{IOI trade-off}
% \label{tab:edge-intervene-sim}
% \end{table}

\paragraph{Discussion: Greedy threshold-based methods may not be suitable for sheaf discovery.}
Interestingly, we observe that the performance of greedy threshold-based methods (ACDC and EAP) is less stable across tasks and, for more complex tasks, especially the PARAREL tasks (Appendix~\ref{app:pararel_results}), these methods reach near-random performance when given the same sparsity budget as DiscoGP. It is possible that greedy threshold-based methods are simply not well suited to any kind of circuit discovery.

Nevertheless, we should take this opportunity to elaborate on the difference between sheaf discovery and other construals of automatic circuit discovery. First and foremost, the two tasks differ in their goals and motivations. Let us revisit the famous example studied by \citet{wangInterpretabilityWildCircuit2022}: ``\exampleg{\small When Mary and John went to the store, John gave a drink to \rule{1em}{0.5pt}}'', where \exampleg{\small Mary} is the correct answer and \exampler{\small John} is the incorrect one. Up to now, the automatic circuit-discovery task has aimed to identify all the important computation edges and components that, when perturbed, cause the greatest change to the final output, potentially steering the model away from responding \exampleg{\small Mary} to \exampler{\small John}. Our results show that simply taking the collection of these important components does not always yield a self-contained mechanism that can perform the task in isolation. Sheaf discovery, on the other hand, aims to capture and identify such a self-contained mechanism (the sheaf).

\begin{table}
\centering\small
\begin{tabular}{@{}lccc@{}}
\toprule
\multicolumn{1}{c}{\multirow{2}{*}{Task}} & \multicolumn{3}{c}{Clean-Ablated Edge similarity}   \\ \cmidrule(l){2-4} 
\multicolumn{1}{c}{}                      & Mean & Interchange & Random \\ \midrule
Agreement & 0.878 & 0.907 & 0.582 \\
IOI       & 0.943 & 0.996 & 0.597 \\
OQA PARAREL & 0.951 & 0.960 & 0.556 \\ \bottomrule
\end{tabular}
\caption{Average cosine similarity between clean and corrupted ablated edge representations across three datasets. Mean and interchange ablations do not substantially affect the models' overall performance.}
\label{tab:edge-intervene-sim}
% \vspace{-1em}
\end{table}

ACDC and EAP should use ablation and they do: both include \emph{mean ablation}, which sets the activation to the average output across a reference distribution obtained by running a sample dataset through the model; and \emph{interchange ablation}, which replaces the activation with its value from corrupted input, created by modifying specific input tokens. These two ablation methods are not suitable for sheaf discovery, as mean-ablated and interchange-ablated components may still retain a large amount of task-related information (Table~\ref{tab:edge-intervene-sim}). This observation is supported by recent work \citep{adolfiComputationalComplexityCircuit2025,shiHypothesisTestingCircuit2024} showing that these ablation- and patching-based methods may not achieve optimal functional fidelity.

\section{Analysis and Findings}
\label{sec:analysis_findings}

% This section enumerates our findings from our analyses and proposes some applications to DiscoGP and sheaf discovery in general.

\begin{table}[!]
    \centering\scriptsize
    \begin{tabular}{l|c c c c c c c c c} \toprule
    \multirow{2}{*}{Task} & \multicolumn{6}{c}{\it Evaluation Tasks} \\
     & AGA & ANA & DNA & DNA i & DNA a & DNA ai \\ \midrule
    AGA & - & 98.0 & 99.7 & 99.7 & 91.9 & 94.8 \\
    ANA & 94.0 & - & 99.7 & 100 & 91.9   & 92.0 \\
    DNA & 92.3 & 86.3 & - & 93.0 & 90.3 & 91.2 \\
    DNA i & 91.3 & 80.3 & 93.7 & - & 94.4 & 93.1 \\
    DNA a &93.0 & 94.6 & 94.2 & 90.5 & - & 94.9 \\
    DNA ia & 91.7 & 90.1 & 92.3 & 94.5 & 94.2 & - \\
    
    \midrule
    Orig. & 99.0 & 100  & 94.7 & 95.3 & 96.0 & 95.7 \\
    \bottomrule
    \end{tabular}
    \caption{Composing sheaves largely preserves functional performance. Each entry shows the performance (accuracy in \%) of a composed circuit (row + column) evaluated on the task associated with the column. For example, the value in column AGA, row ANA shows the performance of the composed circuit (ANA + AGA) on the AGA task. Original (non-composed) sheaf performance is listed in the final row for reference.}
    \label{tab:composition}
    % \vspace{-1em}
\end{table}

\paragraph{Finding 1: Sheaves identified by DiscoGP can be composed while preserving functionality.}

We find that \emph{functional composition of sheaves} is possible under the DiscoGP framework. That is, suppose we have two sheaves that perform tasks A and B, respectively. Simply composing their masks, $m = m_A \cup m_B$, can yield a new sheaf that performs both tasks with largely the same performance. Table~\ref{tab:composition} shows the performance of such compositions across different BLiMP paradigms.

Overall, we observe good composition performance, with the composed sheaves' accuracies reaching 80-100\% across all BLiMP paradigms. To the best of our knowledge, our result is the first successful sheaf or circuit composition in the wild. \citet{mondorfCircuitCompositionsExploring2025} studied circuit composition, but their experiments were limited to synthetic toy models generated using Tracr \citep{lindnerTracrCompiledTransformers2023}.
Composition would be important for demonstrating the practical utility of sheaves because it means that primitive sheaves can be combined to perform multiple, or possibly more complex tasks without having to re-discover every possible combination.

This suggests that at least some degree of modularity is salvageable from LMs after pre-training, although clearly more experimentation is necessary. We hope this finding will motivate future work on modularity and sheaf composition.

% We next conduct several fine-grained analyses of individual circuits discovered by DiscoGP to reveal some of their desired properties that existing circuit discovery methods do not enjoy.
\begin{table}[!]
\small\centering
\setlength\tabcolsep{4pt}
\begin{tabular}{llcc}\toprule
Sheaf 1 & Sheaf 2 & Edge Overlap & Weight Overlap \\
\midrule
AGA & DNA & 14.86\% (251) & 2.69\% (8020) \\
ANA & DNA & 16.19\% (277) & 1.12\% (14816) \\ \midrule
ANA & AGA & 18.32\% (266) & 0.91\% (17693) \\
DNA & DNA irr & 21.07\% (317) & 4.72\% (69364) \\
DNA & DNA adj & 18.46\% (332) & 4.96\% (74782) \\
DNA & DNA irr adj & 18.24\% (323) & 6.06\% (96727) \\
\bottomrule
\end{tabular}
\caption{Sheaf overlap across different BLiMP tasks. The results indicate a trend where similar tasks exhibit higher sheaf overlaps. The overlap percentages are followed by the exact number of overlaps in brackets.}
\label{tab:overlap}
% \vspace{-1em}
\end{table}

\paragraph{Finding 2: Sheaf similarity reflects functional similarity.}
Table \ref{tab:overlap} illustrates the overlap levels between different sheaves. The overlap percentages are calculated by dividing the number of overlap cases by the size of the logical union of the two masks. In this analysis, we only considered the agreement tasks as their task similarity is easier to perceive. BLiMP offers several variants of the DNA pardigms, and we observe here a relatively high level of sheaf overlap in terms of both weights and edges. ANA and AGA, on the other hand, exhibit greater similarity to each other than to DNA as paradigms, because ANA and AGA follow similar sentence templates (see Appendix \ref{app:data}). This similarity between ANA and AGA is reflected in the degree of edge overlap, but not in weight overlap. We conjecture that this distinction between weight and edge overlap is due to the different roles they may play: weights store information, while edges guide the function of the task. While ANA and AGA share similar templates (and therefore exhibit higher edge overlap), performing the task requires distinct parametrized information (resulting in lower weight overlap).

\paragraph{Finding 3: Unveiling the factual recall pipeline in GPT.}

\begin{figure}
    % \centering
    \includegraphics[width=\linewidth]{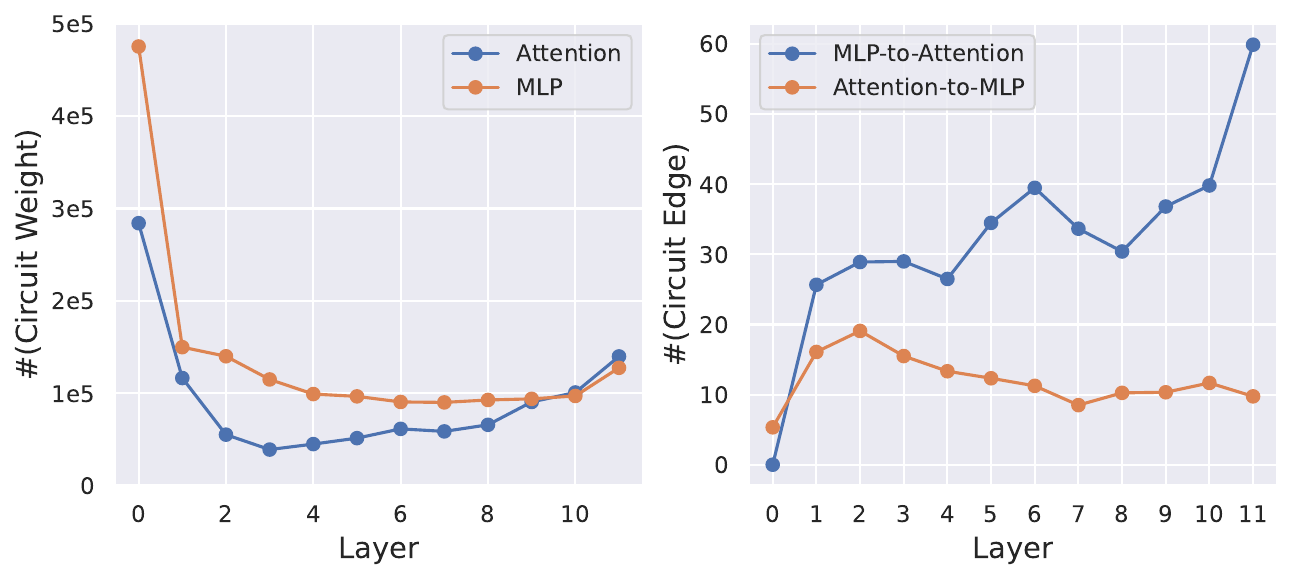}
    % \vspace{-2em}
    \caption{\textbf{Left}: Number of unmasked MLP and attention weights at each layer of the capital city OQA sheaf. \textbf{Right}: Number of edges ending at each layer from preceding MLPs to current-layer attention heads and from preceding attention heads to current-layer MLPs. }
    \label{fig:oqa-circuit-analysis}
    % \vspace{-1em}
\end{figure}

Lastly, we find some confirmation of the \emph{factual recall pipeline} hypothesis, namely that recall occurs in two distinct stages \citep{mengLocatingEditingFactual2022,gevaDissectingRecallFactual2023,niuWhatDoesKnowledge2024,hernandezLinearityRelationDecoding2024}. The left panel of Figure~\ref{fig:oqa-circuit-analysis}
illustrates the layer-wise average number of MLP and attention weight parameters retained in the 12 relation-specific DiscoGP sheaves learned from PARAREL. We observe that MLPs retain substantially more weights in the OQA sheaves compared to attention heads, especially in the lower transformer layers. This finding aligns with recent work that has observed that MLP sublayers function as key-value memory for factual knowledge \textit{extraction} \cite{gevaTransformerFeedForwardLayers2022}. Conversely, the right panel of Figure~\ref{fig:oqa-circuit-analysis} shows the number of sheaf edges at each layer, detailing connections from lower-layer attention heads to current-layer MLPs (Attention to MLP) and from preceding MLPs to current-layer attention heads (MLP to Attention). Notably, the set of connections in upper layers is dominated by MLP-to-attention edges. This observation supports recent findings in mechanistic interpretability suggesting that attention heads play a major role in \textit{propagating} the retrieved factual knowledge from early-site MLPs to upper transformer layers, thereby selecting the most relevant information for answering questions \citep{gevaDissectingRecallFactual2023}.

% Analysis points:
% - The existence of competing circuits

\section{Conclusion}
\label{sec:conclusion}

In this work, we have proposed the notion of a sheaf, and a novel means of discovering them, DiscoGP.  The sheaves discovered by DiscoGP have high functional fidelity using few connections and edges, by combining weight and edge pruning. This method operates with neuron-level granularity and reveals several novel insights into the internal workings of LMs (sheaf modularity and overlap), while also confirming a previously observed trend (the factual recall pipeline).

% Through a comprehensive re-evaluation of prior research on circuit discovery, we have pinpointed their significant shortcomings: the inability to provide both functionally accurate and structurally simple explanations for LM capabilities. To address these deficiencies, we propose sheaf discovery as a generalization of circuit discovery to analyze LM working mechanism, and introduce DiscoGP, an innovative differentiable algorithm that performs joint weight and connection pruning of neural network computation graph to distill fully functional sheaf structures. DiscoGP outperforms existing circuit discovery methods by uncovering more faithful and sparse LM subnetwork structures on multiple NLP tasks. Our analyses showcase how DiscoGP paves the way for novel avenues of language model interpretability, thereby enriching our understanding of the inner workings of powerful yet black-boxed state-of-the-art generative AI systems.

% \newpage

\section*{Acknowledgement}
We thank the anonymous reviewers and the area chair for their insightful and constructive feedback. 

\section*{Limitations}
While our experimental setup is sufficiently comprehensive for the purposes of this study, there is always room to expand the range of tasks and language models evaluated. We focus on GPT-2 to enable direct comparisons with other publicly available systems, but future work could consider larger or more recent models. Additionally, our experiments are limited to English, and extending the analysis to other languages would help assess the generality of our findings.

% Bibliography entries for the entire Anthology, followed by custom entries
%\bibliography{anthology,custom}
% Custom bibliography entries only
\bibliography{arxiv}

\appendix
\section{DiscoGP Implementation Details}
\label{app:implementation_details}

\paragraph{Post-hoc Sheaf Pruning}
Since the training objective (\ref{eq:DiscoGP-obj}) does not consider graph connectivity, we can further simplify the model by (1) removing a node $v$ from the computation graph if all of its weights have been pruned, and (2) performing a reverse BFS from the output node to eliminate edges that do not contribute to the final result.

\begin{figure}[b]
    \centering
    \includegraphics[width=0.5\linewidth]{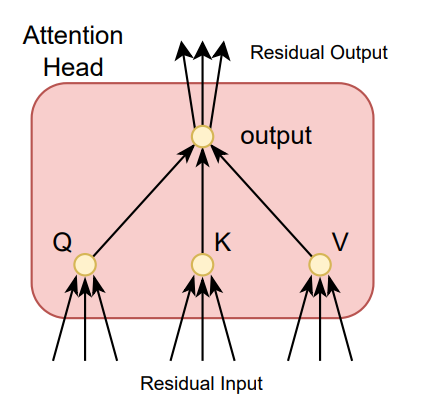}
    \caption{Split QKV Pruning.}
    \label{fig:split-qkv}
\end{figure}

\paragraph{Split QKV Pruning}

Following \citet{conmyAutomatedCircuitDiscovery2023}, we separate the query (Q), key (K) and value (V) activations and introduce an ``output'' node within each attention head. Figure~\ref{fig:split-qkv} shows an illustration of the configuration.

\section{Evaluation Tasks \& Data}
\label{app:data}

\begin{table*}

\scriptsize\centering
\begin{tabular}{@{}lp{2.5cm}p{2.5cm}p{2.5cm}ll@{}}
\toprule
\multicolumn{1}{c}{Agreement Phenomenon} & \multicolumn{1}{c}{Good sentence} & \multicolumn{1}{c}{Bad sentence} & \multicolumn{1}{c}{Converted input query} & \multicolumn{1}{c}{True answer} & \multicolumn{1}{c}{False answer} \\ \midrule
Anaphor Gender Agreement                 & Katherine can't help herself.     & Katherine can't help himself.    & Katherine can't help                      & herself                         & himself                          \\
Anaphor Number Agreement                 & Susan revealed herself.           & Susan revealed themselves.       & Susan revealed                            & herself                         & themselves                       \\ 
\midrule
Det Noun Agr. 1 & Raymond is selling this sketch. & Raymond is selling this sketches. & Raymond is selling this & sketch & sketches \\
Det Noun Agr. Irr. 1 & Laurie hasn't lifted those cacti. & Laurie hasn't lifted those cactus. & Laurie hasn't lifted those & cacti & cactus \\
Det Noun Agr. with Adj. 1 & Rebecca was criticizing those good documentaries. & Rebecca was criticizing those good documentary. & Rebecca was criticizing those good & documentaries & documentary \\
Det Noun Agr. with Adj. Irr. 1 & Some waiters broke this lost foot. & Some waiters broke this lost feet. & Some waiters broke this lost & foot & feet \\
\bottomrule
\end{tabular}%
\caption{Examples of the BLiMP and their converted data.}
\label{tab:anaphor-agreement-samples}
\end{table*}

\begin{table*}
\centering\small
\begin{tabular}{@{}l@{}}
\toprule
\multicolumn{1}{c}{Templates}                                                                                          \\ \midrule
Then, {[}B{]} and {[}A{]} went to the {[}PLACE{]}. {[}B{]} gave a {[}OBJECT{]} to {[}A{]}                                \\ \midrule
Then, {[}B{]} and {[}A{]} had a lot of fun at the {[}PLACE{]}. {[}B{]} gave a {[}OBJECT{]} to {[}A{]}                    \\ \midrule
Then, {[}B{]} and {[}A{]} were working at the {[}PLACE{]}. {[}B{]} decided to give a {[}OBJECT{]} to {[}A{]}             \\ \midrule
Then, {[}B{]} and {[}A{]} were thinking about going to the {[}PLACE{]}. {[}B{]} wanted to give a {[}OBJECT{]} to {[}A{]} \\ \midrule
Then, {[}B{]} and {[}A{]} had a long argument, and afterwards {[}B{]} said to {[}A{]}                                    \\ \midrule
After {[}B{]} and {[}A{]} went to the {[}PLACE{]}, {[}B{]} gave a {[}OBJECT{]} to {[}A{]}                                \\ \midrule
When {[}B{]} and {[}A{]} got a {[}OBJECT{]} at the {[}PLACE{]}, {[}B{]} decided to give it to {[}A{]}                    \\ \midrule
When {[}B{]} and {[}A{]} got a {[}OBJECT{]} at the {[}PLACE{]}, {[}B{]} decided to give the {[}OBJECT{]} to {[}A{]}      \\ \midrule
While {[}B{]} and {[}A{]} were working at the {[}PLACE{]}, {[}B{]} gave a {[}OBJECT{]} to {[}A{]}                        \\ \midrule
While {[}B{]} and {[}A{]} were commuting to the {[}PLACE{]}, {[}B{]} gave a {[}OBJECT{]} to {[}A{]}                      \\ \midrule
After the lunch, {[}B{]} and {[}A{]} went to the {[}PLACE{]}. {[}B{]} gave a {[}OBJECT{]} to {[}A{]}                     \\ \midrule
Afterwards, {[}B{]} and {[}A{]} went to the {[}PLACE{]}. {[}B{]} gave a {[}OBJECT{]} to {[}A{]}                          \\ \midrule
Then, {[}B{]} and {[}A{]} had a long argument. Afterwards {[}B{]} said to {[}A{]}                                        \\ \midrule
The {[}PLACE{]} {[}B{]} and {[}A{]} went to had a {[}OBJECT{]}. {[}B{]} gave it to {[}A{]}                               \\ \midrule
Friends {[}B{]} and {[}A{]} found a {[}OBJECT{]} at the {[}PLACE{]}. {[}B{]} gave it to {[}A{]}                          \\ \bottomrule
\end{tabular}%
\caption{Sentence templates for generating the IOI dataset.}
\label{table:ioi-templates}
\end{table*}

\begin{table*}
\centering\small
\begin{tabular}{p{3cm}p{8cm}}
\toprule
Placeholder Type & Candidate Infilling Words \\ \midrule
{[}A{]} and {[}B{]} (names) & Michael, Christopher, Jessica, Matthew, Ashley, Jennifer, Joshua
Daniel, David, James, Robert, John, Joseph, Andrew, Ryan, Bran
Justin, Sarah, William, Jonathan, Stephanie, Brian, Nicole, Nicho
Heather, Eric, Elizabeth, Adam, Megan, Melissa, Kevin, Steven,
Timothy, Christina, Kyle, Rachel, Laura, Lauren, Amber, Brittan
Richard, Kimberly, Jeffrey, Amy, Crystal, Michelle, Tiffany, Jere
Mark, Emily, Aaron, Charles, Rebecca, Jacob, Stephen, Patrick,
Kelly, Samantha, Nathan, Sara, Dustin, Paul, Angela, Tyler, Scot
Andrea, Gregory, Erica, Mary, Travis, Lisa, Kenneth, Bryan, Lin
Jose, Alexander, Jesse, Katie, Lindsay, Shannon, Vanessa, Court
Alicia, Cody, Allison, Bradley, Samuel. \\ \midrule
{[}PLACE{]} & store, garden, restaurant, school, hospital, office, house, station. \\ \midrule
{[}OBJECT{]} & ring, kiss, bone, basketball, computer, necklace, drink, snack. \\
\bottomrule
\end{tabular}%
\caption{Candidate infilling words of IOI sentence templates.}
\label{table:ioi-infill-words}
\end{table*}

\begin{table*}[]
    \centering\small
    \setlength\tabcolsep{3pt}
\begin{tabular}{ccccc}
\toprule
Relation ID & Relation & No. of Queries & Sample Query & True answer \\ \midrule
P103                            & native language              & 977                                & The mother tongue of Victor Horta is                & Dutch                           \\
P138                            & named after                  & 645                                & Rawlings Gold Glove Award, which is named for       & glove                           \\
P159                            & headquarters location        & 967                                & The headquarter of Strait Shipping is located in    & Wellington                      \\
P176                            & manufacturer                 & 982                                & Honda RA272 is produced by                          & Honda                           \\
P264                            & record label                 & 429                                & Johnny Carroll's record label is                    & Decca                           \\
P279                            & subclass of                  & 964                                & Nucleoprotein 62, a type of                           & protein                         \\
P30                             & continent                    & 975                                & Romulus Glacier is located in                       & Antarctica                      \\
P407                            & language of work or name     & 877                                & Ten Years Gone is a work written in                 & English                         \\
P449                            & original network             & 881                                & Himalaya with Michael Palin was originally aired on & BBC                             \\
P495                            & country of origin            & 909                                & Mundo Obrero was from                               & Spain                           \\
P1376                           & capital of                   & 234                                & Guangzhou is the capital of                         & Guangdong                       \\
P36                             & capital                      & 703                                & The capital city of Porto District is               & Porto                           \\ \bottomrule
\end{tabular}%
% }
\caption{PARAREL relations and sample queries used for circuit discovery.}

\label{tab:pararel-relations}
\end{table*}

\paragraph{BLiMP} BLiMP \citep{warstadtBLiMPBenchmarkLinguistic2020} consists of 67 individual datasets, each containing minimally different sentence pairs that contrast in grammatical acceptability and isolate specific phenomena in syntax, morphology, or semantics. However, BLiMP was designed for bidirectional LMs such as BERT, which require the model to attend to both preceding and following context. Therefore, we use the six BLiMP paradigms applicable to decoder-only LMs (specifically GPT-2). See Table~\ref{tab:anaphor-agreement-samples} for example contrasting sentence pairs and their corresponding query prompts for circuit discovery.

\paragraph{Indirect object identification} \citet{wangInterpretabilityWildCircuit2022} created dataset samples for IOI using templates with random single-token names, places and items. We follow their data curation pipeline by taking the same set of 15 templates and candidate-infilling words to generate our sheaf-discovery dataset. At each trial, we randomly draw a template and a set of infilling tokens to construct a full sentence. We then convert the generated sentence into a binary classification question, where the input prompt is the sentence prefix without the last indirect object, and the two candidates for next token are the indirect object and the subject tokens. See Table \ref{table:ioi-templates} and \ref{table:ioi-infill-words} for a complete list of IOI sentence templates and candidate-infilling words.

\paragraph{PARAREL}
We use the PARAREL dataset by \citet{elazarMeasuringImprovingConsistency2021}, which consists of 38 relation types and 27,738 (subject, relation, object) fact triples such as (\textit{Canada}, capital city, \textit{Ottawa}). We then use the templates created by \citep{daiKnowledgeNeuronsPretrained2022} to convert each fact triple into multiple query prompts (e.g. ``\exampleg{\small The capital city of Canada is \rule{1em}{0.5pt}}''). We take prompts generated from triples with 12 out of 38 PARAREL relations that satisfy the following two conditions: 1) there is a unique object entity answer for each (subject, relation) pair; and 2) the object word always comes at the end of the template-generated sentence so that it can be predicted by an autoregressive language model. We finally obtained a total of 9,543 queries as our dataset of open-domain question answering, and we learn a circuit for each relational dataset for every circuit-discovery method.  See Table \ref{tab:pararel-relations} for a list of the 12 relations we used together with the example fact triples and queries. 

\begin{table*}
    \centering\scriptsize
    \setlength\tabcolsep{3pt}
    \resizebox{\linewidth}{!}{
    \begin{tabular}{l|c|ccc|ccc}
    \toprule

    \multirow{2}{*}{Task}
    
    & Discovery & Test Set Acc. (\%) & Train Set Acc. (\%) & KL Div. & Comp. Acc. (\%) & Weight Density (\%) & Edge Density (\%) \\
    & Method & {\scriptsize (higher is better)} & {\scriptsize (higher is better)} & {\scriptsize (lower is better)} & {\scriptsize (random$^*$ is better)} & {\scriptsize (lower is better)} & {\scriptsize (lower is better)} \\ \midrule

          & ACDC & 0.30 & 0.27 & 0.3194 & 1.20 & 100 & 4.57 \\
          & EAP & 1.18 & 1.63 & 0.3900 & 0.08 & 100 & 6.42 \\
    P30   & Edge    & 92.1 & 89.5  & 0.0115 & 0.90 & 100  & \textbf{2.34} \\
          & Weight  & 86.8 & 92.6  & 0.0093 & 0.23 & 3.86 & 100  \\
          & DiscoGP & \textbf{95.6} & \textbf{92.6}  & \textbf{0.0076} & 0.35 & \textbf{3.64} & 3.01 \\ \midrule

          & ACDC & 0.72 & 0.86 & 0.3706 & 0.42 & 100 & 5.99 \\
          & EAP & 1.18 & 1.86 & 0.3272 & 1.21 & 100 & 4.59 \\
    P36   & Edge    & 62.7 & 90.5  & 0.0164 & 0.86 & 100  & 3.45 \\
          & Weight  & 67.3 & 90.3  & 0.0191 & 1.04 & 4.54 & 100  \\
          & DiscoGP & \textbf{69.2} & \textbf{91.1}  & \textbf{0.0094} & 0.85 & \textbf{4.17} & \textbf{3.22} \\ \midrule

          & ACDC & 0.54 & 1.16 & 0.2913 & 0.36 & 100 & 5.18 \\
          & EAP & 0.93 & 0.57 & 0.3329 & 0.51 & 100 & 5.32 \\
    P103  & Edge    & 91.4 & 88.1  & 0.0345 & 0.88 & 100  & 2.02 \\
          & Weight  & 83.0 & 87.4  & 0.0231 & 0.96 & \textbf{4.35} & 100  \\
          & DiscoGP & \textbf{93.5} & \textbf{89.7}  & \textbf{0.0202} & 0.15 & 4.7  & \textbf{3.36} \\ \midrule

          & ACDC & 0.96 & 0.59 & 0.3096 & 1.29 & 100 & 4.99 \\
          & EAP & 1.98 & 0.78 & 0.2429 & 0.31 & 100 & 5.40 \\
    P138  & Edge    & 64.9 & 96    & 0.022  & 1.52 & 100  & 2.33 \\
          & Weight  & 63.3 & 92.4  & 0.0375 & 0.73 & 1.57 & 100  \\
          & DiscoGP & \textbf{68.0} & \textbf{94.9}  & \textbf{0.029}  & 0.46 & \textbf{1.34} & \textbf{1.9}  \\ \midrule

          & ACDC & 0.56 & 1.64 & 0.3630 & 0.35 & 100 & 4.92 \\
          & EAP & 1.78 & 1.44 & 0.3011 & 0.30 & 100 & 6.41 \\
    P159  & Edge    & 57.3 & 84.2  & 0.0552 & 0.91 & 100  & \textbf{2.05} \\
          & Weight  & 58.8 & 88.7  & 0.0276 & 0.59 & 3.38 & 100  \\
          & DiscoGP & \textbf{62.5} & \textbf{89.8}  & 0.0168 & 0.57 & \textbf{3.79} & 2.81 \\ \midrule

          & ACDC & 0.53 & 1.77 & 0.3823 & 0.48 & 100 & 6.99 \\
          & EAP & 0.91 & 1.39 & 0.3050 & 1.26 & 100 & 4.89 \\
    P176  & Edge    & 86.5 & 98.6  & 0.0117 & 0.47 & 100  & 3.04 \\
          & Weight  & 86.0 & 99.2  & \textbf{0.0095} & 0.88 & 1.34 & 100  \\
          & DiscoGP & \textbf{95.6} & \textbf{99.4}  & 0.0104 & 0.85 & \textbf{1.01} & \textbf{2.73} \\ \midrule

          & ACDC & 1.51 & 0.51 & 0.2250 & 0.57 & 100 & 4.48 \\
          & EAP & 0.27 & 0.39 & 0.2165 & 1.26 & 100 & 6.24 \\
    P264  & Edge    & 77.3 & 89.4  & 0.0297 & 0.16 & 100  & 2.45 \\
          & Weight  & 82.3 & 90.8  & 0.0266 & 1.24 & 3.58 & 100  \\
          & DiscoGP & \textbf{82.9} & \textbf{90.3}  & \textbf{0.0245} & 0.77 & \textbf{3.36} & \textbf{2.43} \\ \midrule

          & ACDC & 1.30 & 0.54 & 0.3590 & 0.77 & 100 & 4.69 \\
          & EAP & 0.74 & 0.55 & 0.3153 & 0.52 & 100 & 6.34 \\
    P279  & Edge    & 69.5 & 87.0  & 0.0562 & 0.68 & 100  & 4.98 \\
          & Weight  & 75.5 & 93.9  & 0.0337 & 0.13 & 2.53 & 100  \\
          & DiscoGP & \textbf{76.9} & \textbf{95.2}  & \textbf{0.0200} & 0.47 & \textbf{2.14} & \textbf{3.57} \\ \midrule

          & ACDC & 1.41 & 1.51 & 0.3492 & 0.32 & 100 & 4.96 \\
          & EAP & 0.49 & 0.66 & 0.2036 & 0.03 & 100 & 5.78 \\
    P407  & Edge    & 80.1 & 93.9  & 0.0085 & 0.55 & 100  & \textbf{2.1}  \\
          & Weight  & 77.0 & 94.1  & 0.0097 & 0.29 & \textbf{1.94} & 100  \\
          & DiscoGP & \textbf{83.3} & \textbf{95.0}  & \textbf{0.0073} & 0.97 & 2.24 & 2.89 \\

    \bottomrule
    \end{tabular}}
    \caption{Sheaf-Discovery Performance Comparison across PARAREL relations. Again, DiscoGP achieves the best performance across all tasks while mostly using the fewest weight parameters and edges. The best-performing methods are highlighted in {\bf bold}. $^*$: For complement sheaf accuracy, successful searches are expected to yield random performance. Therefore, scores in the vicinity of random indicate good performance, and direct comparison of complement scores is not meaningful. Table continues in Table~\ref{table:oqa-full-2}.}
    \label{table:oqa-full}
\end{table*}
\begin{table*}
    \centering\scriptsize
    \setlength\tabcolsep{3pt}
    \resizebox{\linewidth}{!}{
    \begin{tabular}{l|c|ccc|ccc}
    \toprule

    \multirow{2}{*}{Task}
    
    & Discovery & Test Set Acc. (\%) & Train Set Acc. (\%) & KL Div. & Comp. Acc. (\%) & Weight Density (\%) & Edge Density (\%) \\
    & Method & {\scriptsize (higher is better)} & {\scriptsize (higher is better)} & {\scriptsize (lower is better)} & {\scriptsize (random$^*$ is better)} & {\scriptsize (lower is better)} & {\scriptsize (lower is better)} \\ \midrule

          & ACDC & 0.59 & 1.20 & 0.5230 & 0.20 & 100 & 6.88 \\
          & EAP & 0.87 & 0.33 & 0.4976 & 0.82 & 100 & 6.87 \\
    P449  & Edge    & 70.4 & 93.3  & \textbf{0.0090} & 0.95 & 100  & \textbf{3.36} \\
          & Weight  & 71.4 & 93.7  & 0.0098 & 1.39 & 2.7  & 100  \\
          & DiscoGP & \textbf{74.7} & \textbf{93.7}  & 0.0099 & 1.09 & \textbf{2.58} & 3.43 \\ \midrule

          & ACDC & 0.22 & 0.22 & 0.5130 & 0.21 & 100 & 4.37 \\
          & EAP & 1.30 & 0.47 & 0.4058 & 0.43 & 100 & 6.12 \\
    P495  & Edge    & 65.8 & 86.1  & 0.115  & 0.76 & 100  & 3.92 \\
          & Weight  & 65.4 & 87.1  & 0.102  & 0.70  & 2.54 & 100  \\
          & DiscoGP & \textbf{70.7} & \textbf{90.3}  & \textbf{0.082}  & 0.63 & \textbf{2.08} & \textbf{2.17} \\ \midrule
          & ACDC & 1.22 & 1.76 & 0.5535 & 0.65 & 100 & 6.14 \\

          & EAP & 0.38 & 0.76 & 0.5551 & 0.40 & 100 & 6.66 \\
    P1376 & Edge    & 49.4 & 89.3  & 0.101  & 0.77 & 100  & 3.57 \\
          & Weight  & 55.2 & 92.5  & 0.082  & 0.24 & \textbf{1.68} & 100  \\
          & DiscoGP & \textbf{57.7} & \textbf{94.6}  & \textbf{0.047} & 0.28 & 2.13 & \textbf{3.36} \\

    \bottomrule
    \end{tabular}}
    \caption{Sheaf-Discovery Performance Comparison across PARAREL relations (Part 2).}
    \label{table:oqa-full-2}
\end{table*}

\section{Baseline Methods}
\label{app:baseline}

For the greedy, threshold-based approaches, we obtain the original implementations released by the authors and adapt them to work with the same task and configurations as DiscoGP.\footnote{ACDC: \url{https://github.com/ArthurConmy/Automatic-Circuit-Discovery/} and EAP \url{https://github.com/Aaquib111/edge-attribution-patching}.} \citet{bhaskarFindingTransformerCircuits2024} have released their implementation online,\footnote{\url{https://github.com/princeton-nlp/Edge-Pruning}} which is equivalent to our edge pruning setting, where no weight pruning is applied.

For the threshold-based greedy search algorithms, since performance is not an objective in the circuit-discovery process, we can obtain circuits with any level of sparsity by adjusting the thresholds. Therefore, we tune the threshold $\tau$ for each task and report the result that has a comparable --- and larger --- sparsity budget than DiscoGP. This puts ACDC and EAP at an advantage compared to DiscoGP in the sparsity--performance trade-off, yet our results show that DiscoGP still outperforms both.

\section{Detailed \textsc{ParaRel} Results}
\label{app:pararel_results}

Table~\ref{table:oqa-full} and \ref{table:oqa-full-2} list our PARAREL results. Again, DiscoGP achieves the best performance across all tasks while mostly using the fewest weight parameters and edges. The PARAREL task differs from the BLiMP and IOI tasks in that test set and training set performance diverge significantly. This is expected, as factual information tends to be more dispersed. For example, \citet{daiKnowledgeNeuronsPretrained2022,niuWhatDoesKnowledge2024} found that each piece of factual information (e.g., Canada's capital is Ottawa) can be attributed to a handful of neurons, while \citet{niuWhatDoesKnowledge2024} found that the entire determiner--noun agreement can be attributed to the same amount of neurons.

\end{document}